\documentclass[letterpaper, 10 pt, journal, twoside]{IEEEtran}

\usepackage{graphics} 
\usepackage{epsfig} 
\usepackage{amsmath} 
\usepackage{bm}
\usepackage{amssymb}  
\usepackage{algorithm, algorithmic}
\usepackage{xcolor}
\usepackage{graphicx,booktabs,array}
\usepackage[hidelinks]{hyperref}
\usepackage{caption}
\usepackage{subcaption}

\usepackage{physics}
\usepackage{fancyheadings}

\newcommand{\Gs}{\bm{\Psi}}
\newcommand{\gs}{\psi}

\newcommand{\ignore}[1]{}

\newcommand{\ba}{\begin{array}}
\newcommand{\ea}{\end{array}}
\newcommand{\bc}{\begin{center}}
\newcommand{\ec}{\end{center}}
\newcommand{\be}{\begin{enumerate}}
\newcommand{\ee}{\end{enumerate}}
\newcommand{\bea}{\begin{eqnarray}}
\newcommand{\eea}{\end{eqnarray}}
\newcommand{\beas}{\begin{eqnarray*}}
\newcommand{\eeas}{\end{eqnarray*}}
\newcommand{\beq}{\begin{equation}}
\newcommand{\eeq}{\end{equation}}
\newcommand{\bfig}{\begin{figure}}
\newcommand{\efig}{\end{figure}}
\newcommand{\bi}{\begin{itemize}}
\newcommand{\ei}{\end{itemize}}
\newcommand{\bpic}{\begin{picture}}
\newcommand{\epic}{\end{picture}}
\newcommand{\btabular}{\begin{tabular}}
\newcommand{\etabular}{\end{tabular}}
\newcommand{\btable}{\begin{table}}
\newcommand{\etable}{\end{table}}

\newcommand{\es}{\vfill
                 \rule[-6mm]{170mm}{0.7mm} \\
                 \redw{{\tiny
		  \hfill S-\theslide}}
                 \end{slide}}

\newcommand{\vecXX}[1]{{\mathbf {#1}}}

\def \hbar {{\bar{h}}}

\def \vecm {{\vecXX{m}}}

\makeatletter
\renewcommand*\env@matrix[1][*\c@MaxMatrixCols c]{%
  \hskip -\arraycolsep
  \let\@ifnextchar\new@ifnextchar
  \array{#1}}
\makeatother

\title{A Distributed Multi-Robot Framework for Exploration, Information Acquisition and Consensus}

\begin{document}

\author{Aalok Patwardhan$^{1}$ and Andrew J. Davison$^{1}$
\thanks{$^{1}$Aalok Patwardhan and Andrew J. Davison are with the Dyson Robotics Lab and the Department of Computing, Imperial College London 
{\tt\footnotesize [a.patwardhan21,a.davison]@imperial.ac.uk}}%
        }
\maketitle
\thispagestyle{fancy}
\pagestyle{fancy}
\fancyhf{}
\fancyhead[LE,LO]{This work has been submitted to the IEEE for possible publication. Copyright may be transferred without notice, after which this version may no longer be accessible.}

\begin{abstract}
The distributed coordination of robot 
teams performing complex tasks is challenging to formulate. The different aspects of a complete task such as local planning for obstacle avoidance, global goal coordination and collaborative mapping are often solved separately, when clearly each of these should influence the others for the most efficient behaviour. In this paper we use the example application of distributed information acquisition as a robot team explores a large space to
show that we can formulate the whole problem as a single factor graph with multiple connected layers representing each aspect. We use Gaussian Belief Propagation (GBP) as the inference mechanism, which permits parallel, on-demand or asynchronous computation for efficiency when different aspects are more or less important. This is the first time that a distributed GBP multi-robot solver has been proven to enable intelligent collaborative behaviour rather than just guiding robots to individual, selfish goals.
We encourage the reader to view our demos at https://tinyurl.com/gbpstack.

\end{abstract}


\section{INTRODUCTION}

Multi-robot teams offer the potential of solving complex tasks beyond the abilities of single robots, and for scalability it is desirable to seek distributed solutions where the robots operate with local computation and peer-to-peer communication rather than centralised control.
The fact that robot teams are not yet widely used in real applications perhaps reflects the many aspects of competence that are needed to achieve whole tasks, and the difficulty in providing distributed, yet connected solutions to these. For instance a robot team tackling an environmental inspection task must be capable of localisation, planning, mapping and coordinated active information acquisition. Each one of these competences on its own is difficult to achieve in a distributed manner, but for the whole task to be performed well they must also influence each other: localisation is needed for planning, which in turn is needed for coordinated information acquisition.

In this work we show that various types of competence for robot teams can be modelled as {\em factor graphs}, and that this formulation allows for straightforward and principled coupling between competencies as layers in a graph stack.

The novelty of our method is to show that the various competencies or problem domains that influence each other can be optimised independently and asynchronously. Robots effectively hold and update a local copy of a global state based on ad-hoc collaboration with other robots, reaching consensus on its true value.

We build on recent work which showed that Gaussian Belief Propagation (GBP) is a powerful general tool for distributed factor graph inference, specifically when applied to multi-robot planning \cite{Patwardhan:GBPPlanner}. 

Although our formulation for such a joint-optimisation problem is general, we demonstrate it here in simulations of multi-robot exploration and information acquisition about a global state. In the first scenario a swarm of robots must search for a target region in the environment. In the second, the robots must collectively explore the whole environment and accurately estimate the global state. Finally we investigate the performance of our algorithm under communication limitations.

In summary, our main contributions are:
\begin{itemize}
    \item The first application of GBP to a multi-robot, joint optimisation problem over multiple competencies where robots must reach a consensus on a global state through local measurements.
    \item Qualitative and quantitative simulations to show that our method outperforms related work in swarm source-seeking, and the scalability and robustness of our algorithm under communication limitations.
\end{itemize}

\begin{figure}[t]
    \centering
    \includegraphics[width=\linewidth]{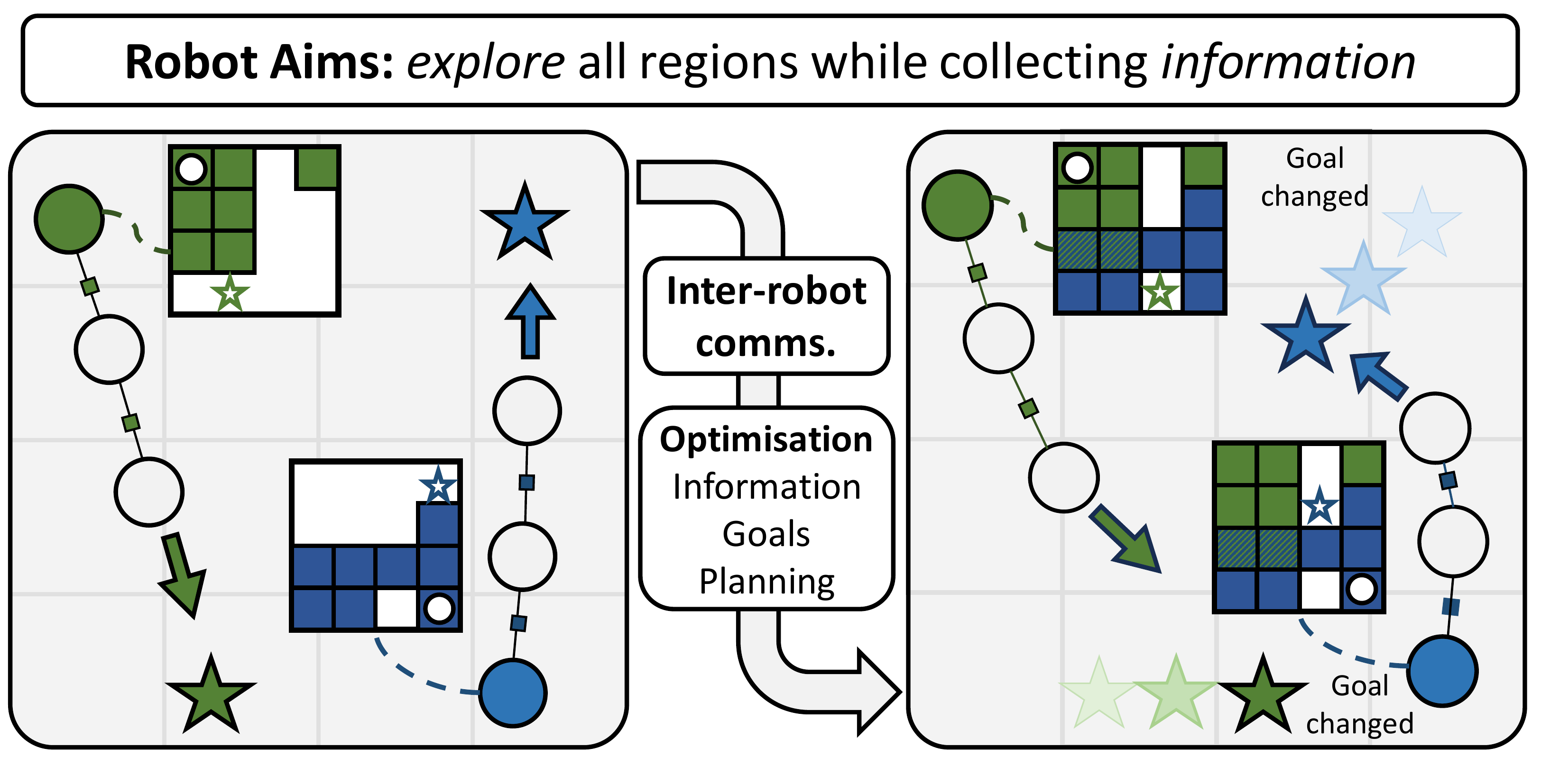}
    \caption{A simplified schematic of the general applications we consider. Left: two robots (circles) have planned their paths towards goals (stars) in unexplored regions based on partial information they have estimated about the global state (shown here as a grid for each robot representing exploration. 
    Right: after message passing they have combined and improved their estimates, optimising their planned paths towards new regions.}
    \label{fig:teaser}
\end{figure}

\section{RELATED WORK}

We are interested in planning for multi-robot teams which have coordinated goals, such as efficient, collaborative information acquisition.
Information-based planning is a challenging problem with wide-ranging applications such as gas distribution mapping \cite{atanasov:gas}, environmental monitoring \cite{woosley:2019}, or target search. Non-myopic information acquisition methods for single robots can be categorised into search-based and sampling-based methods. Search-based methods can find globally optimal paths by searching the motion and information spaces at the expense of computational efficiency \cite{pappas:trajopt}. Sampling-based methods however are able to find solutions, but relax guarantees about optimality by considering subsets of the search space. The centralised algorithm in \cite{kantaros:aia} explores the motion and information spaces by incrementally creating a directed tree, biasing sampling towards informative regions of the environment.

Distributed solutions for multiple robots offer the clear potential for much faster and more scalable information acquisition, but require coordination and communication. The semi-distributed algorithm for gas-plume tracking in \cite{kapoutsis:uav} adjusts a swarm formation to maximise the combined perception of the swarm around areas of high gas concentrations. A centralised unit aggregates measurements made by robots and assigns tasks to each of them to maximise the combined perception of the plume around areas of high gas concentration. 
If global communication between robots can be assumed, algorithms based on Particle Swarm Optimisation (PSO) can be used which model robots as particles performing random walks biased towards the collective best observation regions. These methods do not guarantee success but have been shown to operate well in real-world experiments \cite{sniffybug} where a swarm of small drones localise a gas source in an unknown cluttered environment by communicating observed gas concentrations with the group. Without the guarantee of global connectivity, efficient information dissemination techniques are needed once a robot has completed its task. The hybrid PSO method in \cite{tebert:hybridPSO} simulated robots exploring a complex environment in the form of Perlin noise, searching for a target region below a given threshold. Once a suitable region was found, robots aimed to maximise dissemination of the information across the group with minimal coordination.

These approaches do not consider robot dynamics or noisy signal measurements in their optimisation and produce unrealistic, jerky paths. 
When there are differences in the measurements by two robots of a global property due to sensor noise, they should be able to reach a consensus on the true value \cite{shirsat:multirobot}. 

Multi-robot planning is a well studied topic in literature where robots must plan paths through an environment while avoiding each other and obstacles which may be dynamic. Local distributed planners optimise for the planned paths of robots in a short forward time window, and allow for efficient re-planning of paths in reaction to changes in the environment through communication with other robots \cite{zhou:egoswarm}, \cite{Luis:etal:RAL2020}.

Recently, Gaussian Belief Propagation (GBP) has been shown to be an ideal tool for distribution of convergent computation between multiple robots, whether for multi-robot localisation~\cite{Murai:etal:ARXIV2022} or short-horizon avoidance planning
\cite{Patwardhan:GBPPlanner} and can offer efficient solutions. However, in that work, the multiple robots all had {\em separate}, selfish goals. In this new paper, for the first time we show that an integrated, multi-layer factor graph approach with GBP can be used to achieve coordinated multi-robot behaviour for information acquisition.  The different elements of the whole system --- local avoidance planning, global information-guided goal coordination and coordinated, convergent map creation, can all be formulated as linked factor graphs.

\section{Technical Background}

\subsection{Factor Graphs}
Factor graphs are a useful tool for representing sparse networks of interrelated variables and constraints.
A factor graph is an undirected bipartite graph and represents the factorisation of a joint function $p(\bm{\mathrm{X}})$ into components $f_s$ which depend on subsets $\bm{\mathrm{X}}_s$ of all variables $\bm{\mathrm{X}}$: 
\beq
\label{eqn:factorprod}
p(\bm{\mathrm{X}}) = \prod_s f_s(\bm{\mathrm{X}}_s)
~.
\eeq

In a Gaussian factor graph all factors have the form of Gaussian distributions:
\beq
\label{equ:generalfactor}
f_s(\bm{\mathrm{X}}_s) \propto e^{-\frac{1}{2} \left[ (\bm{z}_s - \bm{h}_s(\bm{\mathrm{X}}_s))^\top \bm{\Lambda}_s (\bm{z}_s - \bm{h}_s(\bm{\mathrm{X}}_s))   \right]  }
~,
\eeq
where $\bm{h}_s(\bm{\mathrm{X}}_s)$ is the functional  form of the constraint that the factor represents, $\bm{z}_s$ is its observed or expected value and $\bm{\Lambda}_s$ is the precision (inverse covariance) of the constraint.

Factor graph inference involves finding the values of variables $\bm{\mathrm{X}}$ which maximise  $p(\bm{\mathrm{X}})$. This is equivalent to finding $\bm{\mathrm{X}}$ to {\em  minimise} the `energy' $E(\bm{\mathrm{X}}) = -\log p(\bm{\mathrm{X}})$. For a Gaussian factor graph, this is effectively minimising a sum of squared terms:
\bea
E(\bm{\mathrm{X}}) = L + \frac{1}{2} \sum_s   (\bm{z}_s - \bm{h}_s(\bm{\mathrm{X}}_s))^\top \bm{\Lambda}_s (\bm{z}_s - \bm{h}_s(\bm{\mathrm{X}}_s))
\label{eqn:factorsum}
~,
\eea
where $L$ is a constant.

Multi-robot optimisation problems that can be formulated as least squares minimisation over the variables can be solved by performing inference on a Gaussian factor graph.
Powerful factor graph inference tools which were originally developed for  probabilistic inference could be used. They have been demonstrated for single robot planning using centralised solvers such as GTSAM \cite{Dellaert:AR2021}, and for multi-robot planning in \cite{Patwardhan:GBPPlanner}.
In this work we use a multi-layered factor graph to represent an abstraction for multi-robot planning problems that are dependent on information acquisition about a global state as illustrated in Figure~\ref{fig:gbpstack}. We will explain the details of the factors shown in the sections that follow.

\subsection{Gaussian Belief Propagation (GBP)}

GBP allows inference on Gaussian factor graphs via distributed computation. Storage of the graph can be shared between multiple devices communicating by radio or some other channel because convergence is achieved via node-wise computation and message passing around the graph.  
GBP is the special case of more general loopy belief propagation and has demonstrated excellent performance, obtaining exact solutions for the marginal means of all variables with rapid convergence~\cite{Ortiz:etal:ARXIV2021}.
GBP has been used \cite{Murai:etal:ARXIV2022} to solve the multi-robot problem of scalable localisation of large numbers of robots making noisy inter-robot measurements.

GBP proceeds by performing iterations of message passing between variables and factors around the loopy factor graph. The current marginal beliefs for variables can be obtained at any time. We now show the main steps; 
see~\cite{Ortiz:etal:ARXIV2021}, \cite{Murai:etal:ARXIV2022}, \cite{Patwardhan:GBPPlanner} for full details.

We represent a Gaussian distribution in canonical/information form as:
\beq
    \mathcal{N}(\bm{\mathrm{X}}; \bm{\mu}, \bm{\Sigma}) = \mathcal{N}^{-1}(\bm{\mathrm{X}}; \bm{\eta}, \bm{\Lambda})~,
\eeq
where  $\bm{\Lambda} = \bm{\Sigma}^{-1}$ and $\bm{\eta} = \bm{\Lambda} \bm{\mu}$ are the precision matrix and information vector respectively.
In GBP, variables $\bm{\mathrm{X}}_s$ are assumed to be Gaussian; thus, each variable $\bm{\mathrm{x}}_k$ has a belief  $b(\bm{\mathrm{x}}_k) = \mathcal{N}^{-1}(\bm{\mathrm{x}}_k; \bm{\eta}_k, \bm{\Lambda}_k)$. 
Factors $F = \{f_s\}_{s=1:N_f}$ are Gaussian constraints between variables; the factor $f_s(\bm{\mathrm{X}}_s)$ is an arbitrary (non-linear) function that connects variables $\bm{\mathrm{X}}_s$.

\subsubsection{Variable Belief Update}
\label{sec:variable_belief_update}
A variable $\bm{\mathrm{x}}_k$ updates its belief by taking the product of all incoming messages from its connected factors:
\beq
b(\bm{\mathrm{x}}_k) = \prod_{f\in n(\bm{\mathrm{x}}_k)} \vecm_{f\rightarrow k}(\bm{\mathrm{x}}_k)~,
\eeq
where $n(\bm{\mathrm{x}}_k) \subseteq F$ is the set of factors that the variable $\bm{\mathrm{x}}_k$ is connected to, and $\bm{\mathrm{m}}_{f\rightarrow k}(\bm{\mathrm{x}}_k) = \mathcal{N}^{-1}(\bm{\mathrm{x}}_k; \bm{\eta}_{f\rightarrow k}, \bm{\Lambda}_{f\rightarrow k})$ is the message from a factor to the variable. In the canonical or information representation of Gaussians, this product can be rewritten as a summation:
\bea 
\bm{\eta}_{k} = \sum_{f\in n(\bm{\mathrm{x}}_k)} \bm{\eta}_{f\rightarrow k}~,~~ \bm{\Lambda}_{k} = \sum_{f\in n(\bm{\mathrm{x}}_k)} \bm{\Lambda}_{f\rightarrow k}~.
\eea


\subsubsection{Variable to Factor Message}
A message from a variable $\bm{\mathrm{x}}_k$ to a factor $f_j \in n(\bm{\mathrm{x}}_k)$ is the product of all incoming factor to variable messages apart from the message from $f_j$:
\beq
\vecm_{\bm{\mathrm{x}}_k\rightarrow j}(f_j) = \prod_{f\in n(\bm{\mathrm{x}}_k) \backslash f_j} \vecm_{f\rightarrow k}(\bm{\mathrm{x}}_k)~.
\eeq

\subsubsection{Factor Likelihood Update}
The likelihood of a factor $f_s(\bm{\mathrm{X}}_s)$ connected to variables $\bm{\mathrm{X}}_s$ with linear measurement function $\bm{h}_s(\bm{\mathrm{X}}_s)$, observation $\bm{z}_s$, and precision $\bm{\Lambda}_s$ can be expressed as a Gaussian distribution $\mathcal{N}^{-1}(\bm{\mathrm{X}}_s; \bm{\eta}_f, \bm{\Lambda}_f)$, where $\bm{\eta}_f = \bm{\Lambda}_s(\bm{\mathrm{z}}_s - \bm{h}_s(\bm{\mathrm{X}}_s))$ and $\bm{\Lambda}_f = \bm{\Lambda}_s$.
If the measurement function is non-linear, we use a first-order Taylor approximation to obtain the likelihood of the linearised factor which takes the form~\cite{Davison:Ortiz:ARXIV2019}:
\bea
    \bm{\eta}_f &=& \bm{J}_s^{\top}\bm{\Lambda}_s\left( \bm{J}_s \bm{\mathrm{X}}_s^0 + \bm{\mathrm{z}}_s - \bm{h}_s(\bm{\mathrm{X}}_s^0)   \right)~, \\
    \bm{\Lambda}_f &=& \bm{J}_s^{\top}\bm{\Lambda}_s \bm{J}_s~,
\eea
where $\bm{J}_s$ is the Jacobian and $\bm{\mathrm{X}}_s^0$ is the linearisation point: the current state of the variables.
In our work $\bm{\mathrm{z}}_s=0$ for all the factors unless stated otherwise, meaning that the factor energy is purely a function of the variables. 

\subsubsection{Factor to Variable Message}
We take the product of the factor likelihood and messages from $\bm{\mathrm{X}}_s \backslash \bm{\mathrm{x}}_k$, then marginalise out all variables but $\bm{\mathrm{x}}_k$ resulting in a message from a factor to variable $\bm{\mathrm{x}}_k$:
\beq
\vecm_{f\rightarrow k}(\bm{\mathrm{x}}_k) = \sum_{\bm{\mathrm{x}} \in \bm{\mathrm{X}}_s \backslash \bm{\mathrm{x}}_k} f_s(\bm{\mathrm{X}}_s) \prod_{\bm{\mathrm{x}} \in \bm{\mathrm{X}}_s \backslash \bm{\mathrm{x}}_k} \vecm_{\bm{\mathrm{x}}\rightarrow f}(\bm{\mathrm{x}})~.
\eeq

\section{METHOD}
We consider the complex problem of exploration and information acquisition where a swarm of robots $\mathcal{R}=[\mathcal{R}_i], i \in \{0..N_R$\} must efficiently explore an environment $\mathcal{M}\in\mathbb{R}^d$ to collaboratively estimate a global state $\Gs$.

In general $\Gs$ can represent any scalar or vector quantity present throughout the environment with values in the range [0,1] of which robots can make noisy and partial measurements. In our work $d=2$ and $\Gs$ represents a scalar signal field such as gas concentration present throughout the environment $\mathcal{M}$ which is segmented into $N_M$ square sampling regions of width $r_D$m. Robots can make measurements of $\Gs$ for regions $m \in \mathcal{M}(r_S)$ within their sampling radius $r_S$.

The aims of the robots are to:
\begin{itemize}
    \item Collaborate with other robots to form a consensus on the global state using noisy local measurements.
    \item Find a location where the signal value is below a given threshold, or collectively explore the whole environment.
    \item Avoid collisions with other robots and plan smooth paths over a time horizon $T_H$ in Euclidean space.
\end{itemize}
A simplified schematic of the problem is shown in Figure \ref{fig:teaser}.
\subsection{A GBP Stack of Factor Graphs}
Robots plan paths towards intermediate goal locations in the environment which directly influence the information they collect about $\Gs$ through sampling.
However, this collected information can also actively guide the path planning goals to new and potentially informative regions; there is a clear inter-dependency between information acquisition, path planning and goal selection.

We formulate the problem as a \emph{stack of factor graphs} which we call the `GBP stack', where the layers of the stack represent different problem domains that can be solved using GBP.
Each robot holds variables and factors in the layers of its own GBP stack, and is able to create and destroy factors between themselves and their neighbours in an ad-hoc manner. In this way optimisation can be done using GBP both \emph{within} a robot's stack as well as \emph{between} a robot and its neighbours.

\begin{figure}[h]
    \centering
    \includegraphics[width=0.8\linewidth]{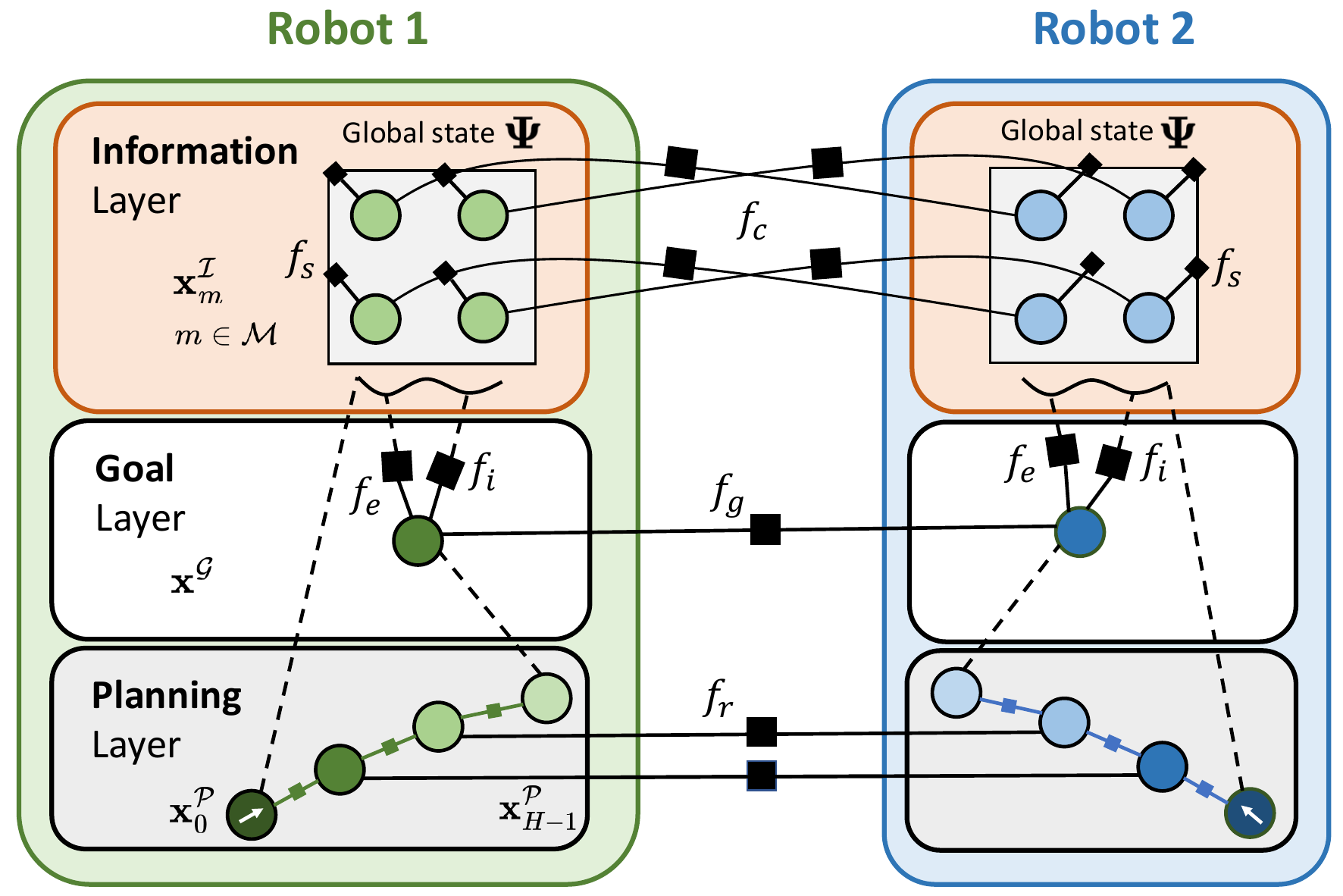}
    \caption{The GBP stack of factor graphs for two robots estimating a global state $\Gs$ shown here as being segmented into $N_M=4$ regions. A robot holds a copy of $\Gs$ in its Information layer which it can update asynchronously. The Goal layer guides the Planning layer for exploration. Message passing can happen asynchronously between the layers of a robot's stack as well the stacks of other robots.}
    \label{fig:gbpstack}
\end{figure}

This is a general formulation, but in this work we consider a robot's GBP stack formed of layers representing information acquisition, goal selection, and local path planning as shown in Figure \ref{fig:gbpstack}. The information layer $\mathcal{I}$ contains a robot's copy of the global state $\Gs$, updated with local measurements and information from other robots. The goal layer $\mathcal{G}$ is responsible for the selection of regions to plan paths towards. Finally the planning layer $\mathcal{P}$ is responsible for the short range path planning and collision avoidance of the robot.

In real-world systems robots may be restricted in their inter-robot communication due to bandwidth or power constraints and may not receive information at every timestep for each layer. The asynchronous nature of GBP allows a robot to continue optimising over $\mathcal{G}$ and $\mathcal{P}$ using the most recent information it holds in $\mathcal{I}$.

We now describe in each layer in detail, and the types of variables and factors that are involved.

\subsection{Information layer $\mathcal{I}$}
\label{sec:info_layer}
This layer represents a robot's copy of the global state; there is one variable $\mathbf{x}_{m}^{\mathcal{I}}$ for each region $m$ in the environment:
\bea
    \mathbf{x}_{m}^{\mathcal{I}} = [\bm{p}_{m}^{\mathcal{I}}, \gs_{m}^{\mathcal{I}}, \zeta_{m}^{\mathcal{I}}]^\top~,
\eea
Dropping the superscript $\mathcal{I}$ for brevity, this variable is formed of the position $\bm{p}_{m}$, signal value ${\gs}_{m}$, and coverage $\zeta_m \in [0,1]$ of the region $m$. $\zeta_m=1$ signifies that the cell has been visited by the robot, and $\gs_m=1$ at initialisation.

\subsubsection{Sensor Factor}
Each variable is connected to a unary factor representing the noisy sensor measurement $\bm{z}_{s}$ made by the robot of the region $m$. The factor is represented as
\bea
\label{eqn:unary}
    \bm{f_s}:~\bm{h}_s(\mathbf{x}_m^{\mathcal{I}}) = \mathbf{x}_m^{\mathcal{I}},~ \bm{\mathrm{\Lambda}}_s = diag(\sigma_p^{-2}, \sigma_{\gs}^{-2}, \sigma_{\zeta}^{-2})~, \\
    \bm{z}_{s} \sim \mathcal{N}([\Tilde{\bm{p}}_{m}, \Tilde{\gs}_{m}, 1]^\top , \bm{\mathrm{\Lambda}}_s^{-1})~.
\eea
where the tilde $\sim$ above a variable represents its true value. In our work $\sigma_p$ and $\sigma_{\zeta}=10^{-5}$, representing perfect measurement of the location of $m$.

\subsubsection{Inter-robot Consensus Factor}
Each variable in this layer is also connected to the corresponding variables in the Information layers of other robots. These factors have precision $\bm{\mathrm{\Lambda}}_c = \sigma_c^{-2}\bm{\mathrm{I}}$ and are represented as
\bea
    \label{eqn:consensusfactor}
    \bm{f_c}:~\bm{h}_c(\mathbf{x}_{m}^{\mathcal{I}_1}, \mathbf{x}_{m}^{\mathcal{I}_2}) = \mathbf{x}_{m}^{\mathcal{I}_1}-\mathbf{x}_{m}^{\mathcal{I}_2}~.
\eea

For a pair of robots $(R_1, R_2)$, these factors encourage $R_1$'s beliefs of the variables to be the same as those of $R_2$, allowing the robots to reach a consensus on the global state $\Gs$ through iterations of GBP, as well as exchanging information about the regions of the environment they have collectively visited.

\subsection{Goal layer $\mathcal{G}$}
\label{sec:GoalLayer}
The Goal layer in a robot's GBP stack is responsible for deciding on regions to plan paths towards whilst avoiding regions that other robots are heading to. Each robot holds a variable in this layer $\mathbf{x}^{\mathcal{G}}$ representing a goal location for the robot's path planning layer. This variable is connected to factors representing information acquisition, exploration and inter-robot coordination.

\subsubsection{Signal Factor}
A factor $\bm{f_i}$ aims to draw the goal variable to the region $m^*$ with the best (lowest) value ${\gs}_{m^*}$ of $\Gs$. It is therefore dependent on the most up-to-date knowledge from the Information layer. The factor precision matrix is $\bm{\mathrm{\Lambda}}_i = \sigma_i^{-2}\bm{\mathrm{I}}$ and the factor is denoted as
\bea
    \label{eqn:f_i}
    \bm{f_i}:~\bm{h}_i(\mathbf{x}^{\mathcal{G}} ; \bm{p}_{m^*}^{\mathcal{I}}) =& u(m^*)( \mathbf{x}^{\mathcal{G}} - \bm{p}_{m^*}^{\mathcal{I}})~, \\
    u(m^*) =& 1-\gs_{m^*}~, \\
    m^* =& \underset{m \in \mathcal{M}}{\rm argmin}~\gs_m~.
\eea

\subsubsection{Exploration Factor}
This factor $\bm{f}_e$ aims to draw the goal variable $\mathbf{x}^{\mathcal{G}}$ towards the nearest unexplored region and is therefore also dependent the Information layer as well the Planning layer with the robot's current position $\bm{p}_{0}^{\mathcal{P}}$. The factor precision is $\bm{\mathrm{\Lambda}}_e = \sigma_e^{-2}\bm{\mathrm{I}}$ and has the same form as (\ref{eqn:f_i}) with
\bea
    \label{eqn:f_e}
    u(m^*) = 1,~~m^* =& \underset{m \in \{\underset{m' \in \mathcal{M}}{\rm argmin}~\zeta_{m'}\}}{\rm argmin}~||\bm{p}_{m^*}^{\mathcal{I}} - \bm{p}_{0}^{\mathcal{P}}||~.
\eea
If there are no unexplored regions, a random location is selected so that robots continue to explore. 
\subsubsection{Goal Diversity Factor}
Finally $\mathbf{x}^{\mathcal{G}}$ is connected to the goal variables of neighbouring robots. This discourages two connected robots from moving towards the same region and therefore promotes exploratory behaviour in the swarm. The factor has precision $\bm{\mathrm{\Lambda}}_g = \sigma_g^{-2}\bm{\mathrm{I}}$ and is formed as
\bea
    \bm{f_g}:~\bm{h}_g(\mathbf{x}^{\mathcal{G}_1}, \mathbf{x}^{\mathcal{G}_2}) = \begin{cases}1 - \frac{\lVert \mathbf{x}^{\mathcal{G}_1} - \mathbf{x}^{\mathcal{G}_2} \rVert}{r_D} & \lVert \mathbf{x}^{\mathcal{G}_1} - \mathbf{x}^{\mathcal{G}_2} \rVert \leq r_D \\
                        0 & \text{otherwise}\end{cases}~.
\eea 

\subsection{Planning layer $\mathcal{P}$}
\label{sec:planning_layer}
This layer is based on our previous work \cite{Patwardhan:GBPPlanner} where a robot's path consisted of variables representing its position and velocity in a short forward time window $T_H$. These variables were connected with factors representing dynamics $\bm{f_d}$ and collision avoidance $\bm{f_r}$, and the planning horizon state had a fixed non-optimisable velocity. 

In this work however the horizon state $\mathbf{x}_{H-1}^{\mathcal{P}}$ is dependent on the Goal layer, having a velocity of magnitude $V_{max}$ towards $\mathbf{x}^{\mathcal{G}}$.
A robot's planned path can therefore dynamically change based on the latest information of the world.

\subsection{Algorithm}
Robots initialise their GBP stacks with the variables and factors in each layer as described in sections \ref{sec:info_layer}~to~\ref{sec:planning_layer}.
At every timestep, each robot follows Algorithm \ref{algo}.

When a new robot $R_j$ is within communication range $r_C$ of robot $R_i$, inter-robot factors are created between the two robots for each layer in the robots' GBP stacks if they do not already exist, and are destroyed when the robots are out of range.

The robot samples the environment for regions $m\in\mathcal{M}(r_S)$ and updates its corresponding Information layer variables $\mathbf{x}_{m}^{\mathcal{I}}$.
Joint optimisation is performed over all layers using GBP for $N_I$ iterations. In the Information layer, only variables representing regions within the robot's communication radius $r_C$ are set to `active' and take part in inter-robot message passing.
Finally the robot's position and horizon states in its planning layer are updated using the simulation timestep $\Delta$T.

\begin{algorithm}[t]
\caption{For each robot $R_i$}\label{algo}
\begin{algorithmic}[1]
 \STATE Initialise GBP stack layers $\mathcal{I,G,P}$
 \WHILE{$t<T_{max}$}
 \STATE \textit{Let $N(R_i) = \{R_j\ |\ ||{\mathbf{p}_0^{\mathcal{P}_i}-\mathbf{p}_0^{\mathcal{P}_j}}|| < r_C  \}$ be the set of robots within the communication radius of $R_i$.}\\
 \STATE \textit{Let $C(R_i)$ be the set of robots connected to $R_i$.}\\
 \FOR {Newly observed robot $R_j \in N(R_i) \backslash C(R_i)$}
 \STATE {Create inter-robot factors $f_c, f_g, f_r$.}
 \ENDFOR
 \FOR {Out-of-range robot $R_j \in C(R_i) \backslash N(R_i)$}
 \STATE {Delete inter-robot factors $f_c, f_g, f_r$.}
 \ENDFOR
 \STATE Make local measurement of $\mathbf{x}_m^{\mathcal{I}}~\forall~m \in \mathcal{M}(r_S)$
 \STATE Activate variables $\mathbf{x}_{m}^{\mathcal{I}}~\forall~m\in \mathcal{M}(r_C)$.
 \FOR {layer $\mathcal{L}$ in GBP Stack}
 \STATE Perform $N_I$ iterations of GBP
 \ENDFOR
 \STATE Update $\mathbf{x}_0^{\mathcal{P}}$ and $\mathbf{x}_{H-1}^{\mathcal{P}}$ by $\Delta T$.
 \ENDWHILE
\end{algorithmic} 
\end{algorithm}

\section{EXPERIMENTS}
We model a swarm of $N_R$ robots with radius $r_R=1$m and planning horizon $T_H=1$s traversing a square environment of length $D$~m, segmented into cells with side length $r_D=10$m.
In the experiments that follow, the robots have a maximum speed of $V_{max}=5$m/s, and make measurements every $1$s within a sampling radius $r_S=10$m.
The communication intervals for the Goal and Information layers of the robots' GBP stacks are set to $T_C^{\{\mathcal{I},\mathcal{G}\}} = 1$s, and $T_C^{\mathcal{P}}=\Delta T=0.1$s. We set $N_I=5$, $\sigma_{c}=1$~[-], $\sigma_e=0.1$ m, $\sigma_i=0.5$m and $\sigma_g=0.01$~[-], encouraging exploration whilst seeking regions of low $\gs$.
 
\subsection{Baseline for Comparison}
We compare our method with the Hybrid Particle Swarm Optimisation (PSO) algorithm of \cite{tebert:hybridPSO}, where a swarm of robots explores a complex environment with a scalar signal field searching for a region with signal value less than a given threshold. Similar to our work, the authors consider a dynamic graph topology of robot connectivity and the method operates in a purely decentralised manner. However it does not account for robot collision avoidance and creates unrealistic jerky paths.
For a fair comparison we use this work as a baseline in the source seeking experiment to show the validity of our method against the same complexities of 2D Perlin noise environments and initial robot configurations. We use the best performing parameters from the work: $c_{p,g}=1, \omega=0$.

\subsection{Source Seeking}
\label{sec:sourceseeking}
Robots begin in one corner of the environment with $D=100$m and search for a region where the value of the global state $\gs$ is less than a threshold $\gs^*=\frac{10}{255}$ (i.e. a `source'). The noise on their measurements of $\gs$ is set to $\sigma_{\gs}=0.01$ for our method while Hybrid PSO assumes perfect sensing.

We measure the time taken for all robots to have found or gained knowledge of the target region as we vary the number of robots $N_R$ and the radius of communication $r_C$. 

\begin{figure}[t]
    \centering
    \includegraphics[width=\linewidth]{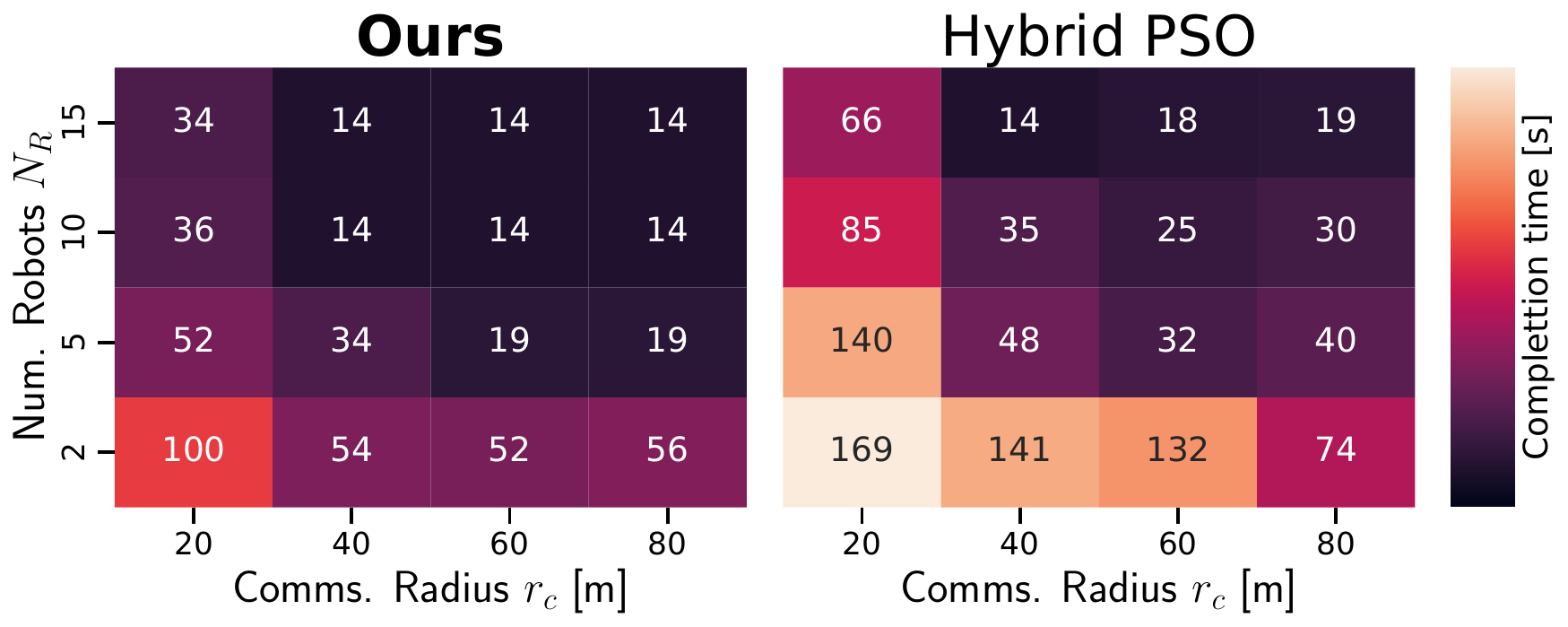}
    \caption{Completion times for the source seeking experiment as the number of robots $N_R$ and communication radius $r_C$ varies. Our method is faster than Hybrid PSO while being able to consider noisy sensor measurements.}
    \label{fig:source_seeking_data}
\end{figure}

\begin{figure}[t]
    \centering
    \includegraphics[width=0.9\linewidth]{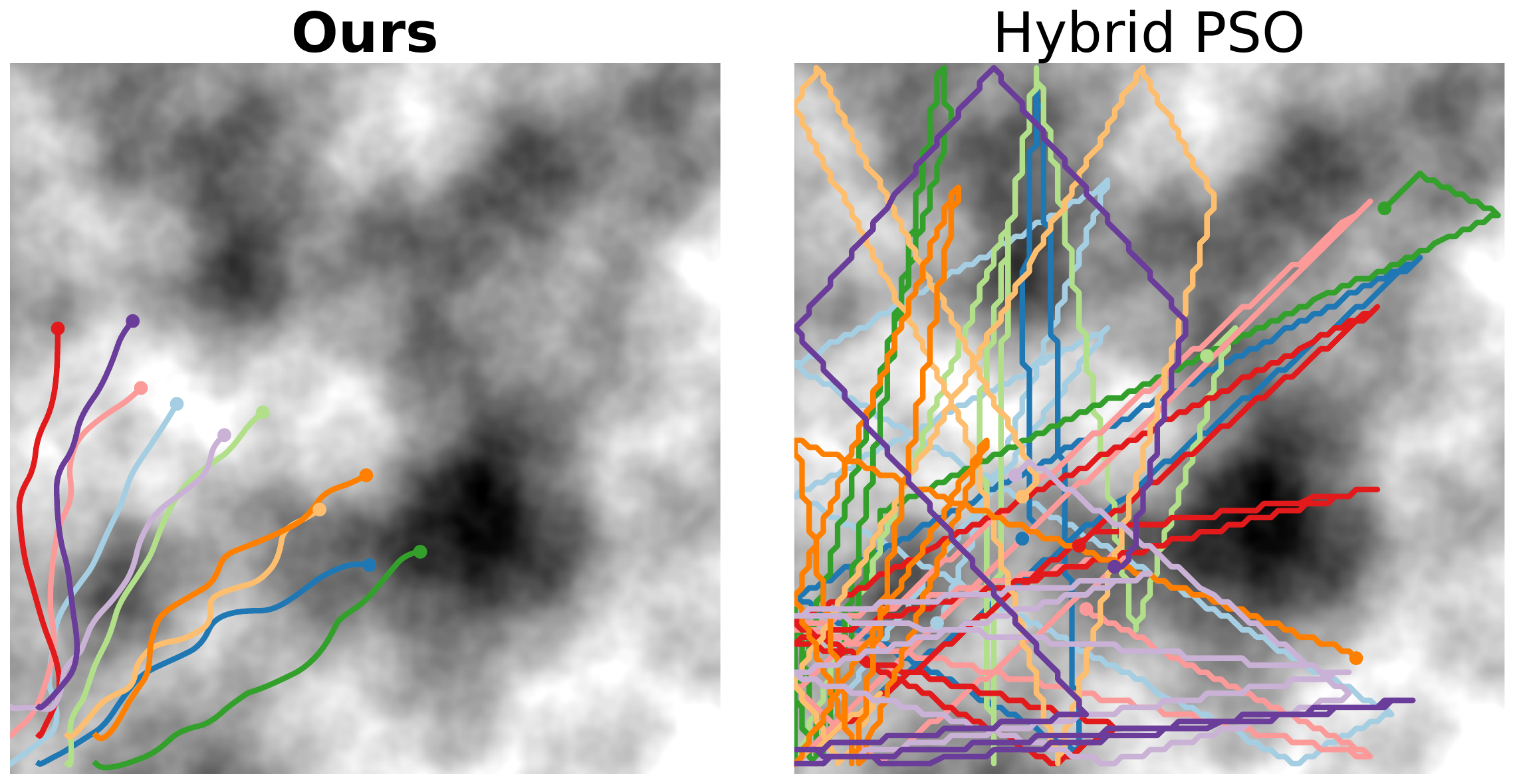}
    \caption{Paths taken until the target region was found in the source seeking experiment for $N_R=10$ robots and communication radius $r_C=40$ m. Our method optimises for smooth paths and efficient coverage of the environment without collisions.}
    \label{fig:source_seeking_paths}
\end{figure}

Figure \ref{fig:source_seeking_data} shows that as $N_R$ increases the completion time decreases as robots quickly spread and cover a larger area of the environment, due to their goal diversity factors $f_g$. This is also seen as $r_C$ increases and robots are able to receive information about more distant regions of the environment through inter-robot communication. A larger $N_R$ results in diminishing returns for completion times due to the increase in robot density in space. Our method outperforms the baseline due to its efficient information sharing and path planning and is able to create smooth paths as seen in Figure \ref{fig:source_seeking_paths}.

\subsection{Exploration and Coverage}
In this experiment, the task of the swarm is to collectively explore an environment of width $D=200$ m, taking noisy local measurements of the global signal field $\Gs$ with $\sigma_{\gs}=0.1$. We weaken the signal factor $\sigma_i=1000$ whilst keeping the other factors of the same strength to encourage exploration. We vary the number of robots $N_R$ in the swarm and consider the evolution of two metrics over time: 

\emph{Coverage $\bm{\zeta}(t)$}: the proportion of the environment visited ($\zeta_m>0$) by a robot, averaged over all robots.

\emph{Global Estimation Error $RMS_{\gs}(t)$}: the root-mean-square difference between a robot's beliefs of $\gs_m$ and the true values $\Tilde{\gs}_{m}$ averaged over all regions $m\in\mathcal{M}$ and all robots.

\begin{figure}[t]
    \centering
    \includegraphics[width=\linewidth]{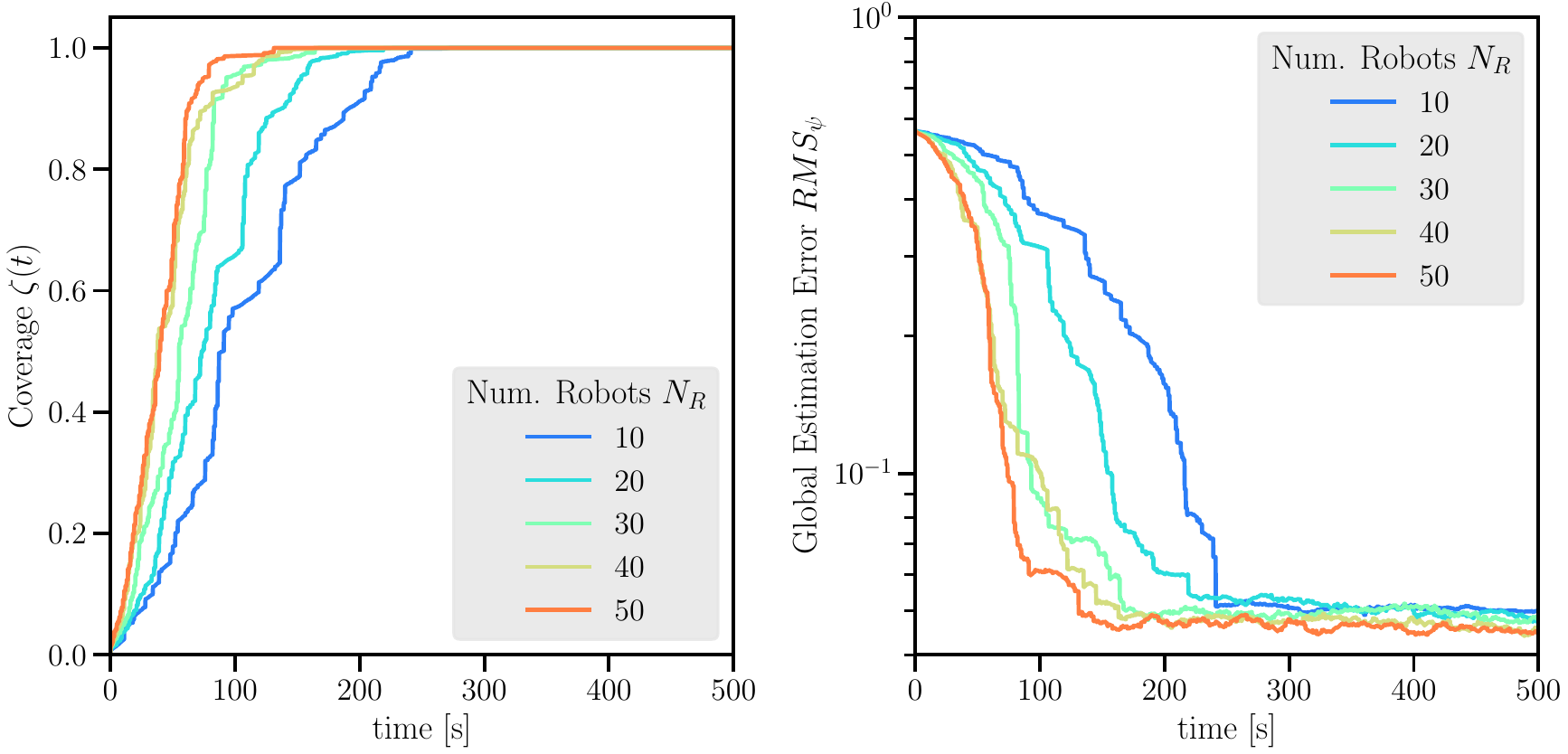}
    \caption{Evolution of coverage and $RMS_{\gs}$ over time for the `random' initialisation scenario. Increasing the number of robots $N_R$ results in quicker coverage of the environment and a steady-state $RMS_{\gs}$ much lower than the measurement noise ($\sigma_s=0.1$) due to greater collaboration.}
    \label{fig:exploration_coverage}
\end{figure}

Robots are initialised in random positions with zero velocity, and Figure \ref{fig:exploration_coverage} shows that increasing the number of robots $N_R$ results in faster coverage; robots perform goal layer optimisation and plan paths towards distinct regions. The steady-state values of $RMS_{\gs}$ decrease as $N_R$ increases, and are significantly lower than the standard deviation of noise $\sigma_{\gs}=0.1$ on robots' sensors. There is clear benefit to collaboration in the swarm, although diminishing returns are seen for both metrics as the density of robots increases.

\subsection{Varying Communications Radius}
When two robots communicate they share information about regions of the environment that are within $r_C$ of their current position. This is useful for scalability as it limits the amount of data transferred between them (robots do not need to share information about their whole map).
For the two initial conditions `corner' and `random' we measure the time taken for full coverage as well as the steady-state value of $RMS_{\gs}$ after $1000$s for representative low, medium and large $r_C$.

Table \ref{table:rc} shows that a large $r_C$ results in faster coverage and lower $RMS_{\gs}$ as robots are able to process information about distant regions and can benefit from being spread out as in the `random' scenario. For small $r_C$, the `corner' scenario (see Figure \ref{fig:source_seeking_paths}) performs better -- robots begin close together resulting in more inter-robot communication as compared to the `random' scenario.

Although a larger communication radius is beneficial for the swarm, this effectively increases the amount of data transferred and presents a trade-off for real-world systems.

\begin{table}[t]
  \caption{Effect of varying communication range $r_C$ for $N_R=20$ in the `corner' and `random' initialisations}
  \label{table:rc}
\centering
\begin{tabular}{r|rr|rr}
\toprule
   & \multicolumn{2}{c|}{Coverage Time [s]} & \multicolumn{2}{c}{Steady-state $RMS_{\gs}$} \\ 
$r_C$ & Corner & Random & Corner    &  Random       \\ \hline
20    & 147   & 232 & 0.0408    &  0.0425         \\
50    & 84   & 56 & 0.0375    &  0.0381         \\
100   & 78   & 48 & 0.0359    &  0.0343         \\
\end{tabular}
\end{table}

\subsection{Performance under Communication Limitations}
A key aspect of our multi-layer approach is that robots do not constantly need to take part in information exchange. Real-world robots may be subject to power limitations or sensor failures, and may prioritise collision avoidance over information acquisition. We simulate this regime of communications failure for $N_R=20$ robots in the `random' initialisation scenario. At every timestep a proportion $\alpha$ of randomly selected robots \emph{do not} take local measurements or perform inter-robot communication in their Goal and Information layers.
\begin{figure}[t]
    \centering
    \includegraphics[width=\linewidth]{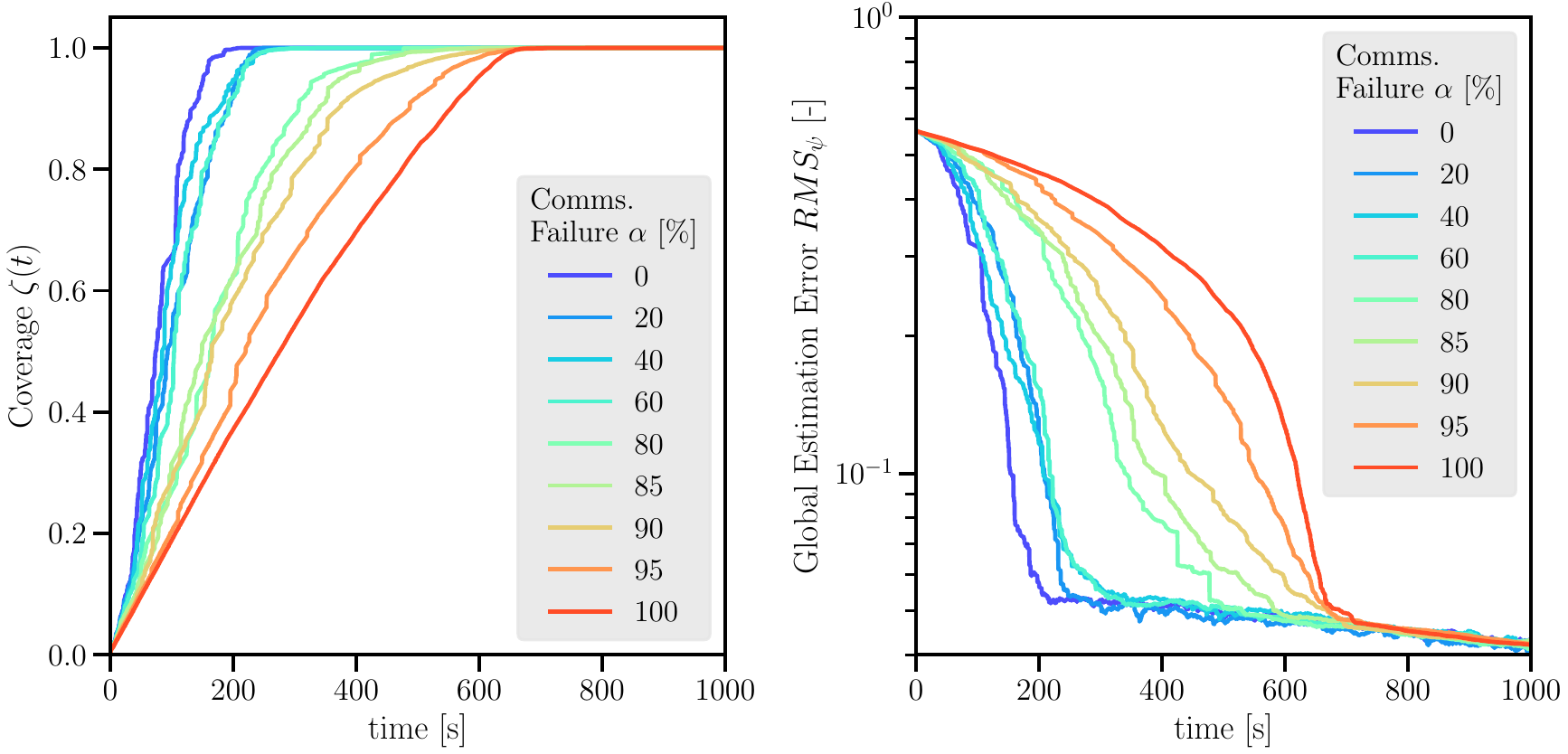}
    \caption{Evolution of coverage and $RMS_{\gs}$ over time as the communications failure rate $\alpha$ varies. Robots take longer to explore the environment as the inter-robot communication decreases although they are still able to accurately estimate the global state.}
    \label{fig:commsfailure_coverage}
\end{figure}

Figure \ref{fig:commsfailure_coverage} shows that as $\alpha$ increases it takes longer for the swarm to cover the full environment and to accurately estimate $\Gs$ as robots cannot efficiently decide on goal locations for path planning, or make use of their neighbours' collected information. 
As $\alpha$ increases robots interact less with their neighbours and continue to optimise over their own GBP stack, making use of new information when it becomes available; full coverage and accurate estimation is still achieved when the robots behave independently ($\alpha=1$).

\section{CONCLUSIONS}
We have shown that our distributed method for collaborative optimisation over multiple problem competencies can enable a team of robots to accurately estimate a global state based purely on local independent measurements. By formulating the complex problem as a stack of linked factor graphs robots are able to optimise over each competence of the problem in an asynchronous manner. The distributed nature of our algorithm which relies on per-robot message passing promises scalability to large environments and numbers of robots.

Our general algorithm is not limited to the square environments considered here as we include the positions of regions in the information layer variables. In the future we will test our algorithm with more complex maps, and investigate optimisation with a dynamic model of the global state as here we have assumed it to be constant with time.

\section*{ACKNOWLEDGEMENTS}

This research has been supported by Dyson Technology Ltd and EPRSC.

\addtolength{\textheight}{-12cm}   





\bibliographystyle{unsrt} 
\bibliography{bib,robotvision}

\end{document}